\documentclass[10pt,twocolumn,letterpaper]{article}

\usepackage{cvpr}
\usepackage{times}
\usepackage{epsfig}
\usepackage{graphicx}
\usepackage{amsmath}
\usepackage{amssymb}
\usepackage{float}

\usepackage{setspace}
\usepackage{textcomp}
\usepackage{url}
\usepackage{multirow}

\usepackage{epsfig}
\usepackage{amsmath}
\usepackage{amssymb}
\usepackage{caption}
\usepackage{subcaption}
\usepackage{sidecap}
\usepackage[dvipsnames]{xcolor}

\DeclareMathOperator*{\argmax}{arg\,max}

\newcommand{\dataset}{VideoGaze}
\cvprfinalcopy

\usepackage[pagebackref=true,breaklinks=true,letterpaper=true,colorlinks,bookmarks=false]{hyperref}

\def\cvprPaperID{354} 

\ifcvprfinal\fi
\begin{document}

\title{Following Gaze Across Views}

\author{Adri\`{a} Recasens\hspace*{6mm}Carl Vondrick\hspace*{6mm} Aditya Khosla\hspace*{6mm}Antonio Torralba\\
Massachusetts Institute of Technology\\
{\tt\small \{recasens,vondrick,khosla,torralba\}@csail.mit.edu}\\
}

\maketitle

\begin{abstract}
      Following the gaze of people inside videos is an important signal for understanding people and their actions. In this paper, we present an approach for following gaze across views by predicting where a particular person is looking throughout a scene. We collect VideoGaze, a new dataset which we use as a benchmark to both train and evaluate models. Given one view with a person in it and a second view of the scene, our model estimates a density for gaze location in the second view. A key aspect of our approach is an end-to-end model that solves the following sub-problems: saliency, gaze pose, and geometric relationships between views. Although our model is supervised only with gaze, we show that the model learns to solve these subproblems automatically without supervision. Experiments suggest that our approach follows gaze better than standard baselines and produces plausible results for everyday situations.
\end{abstract}

\section{Introduction}
\begin{figure}[t!]
\includegraphics[width=0.48\textwidth]{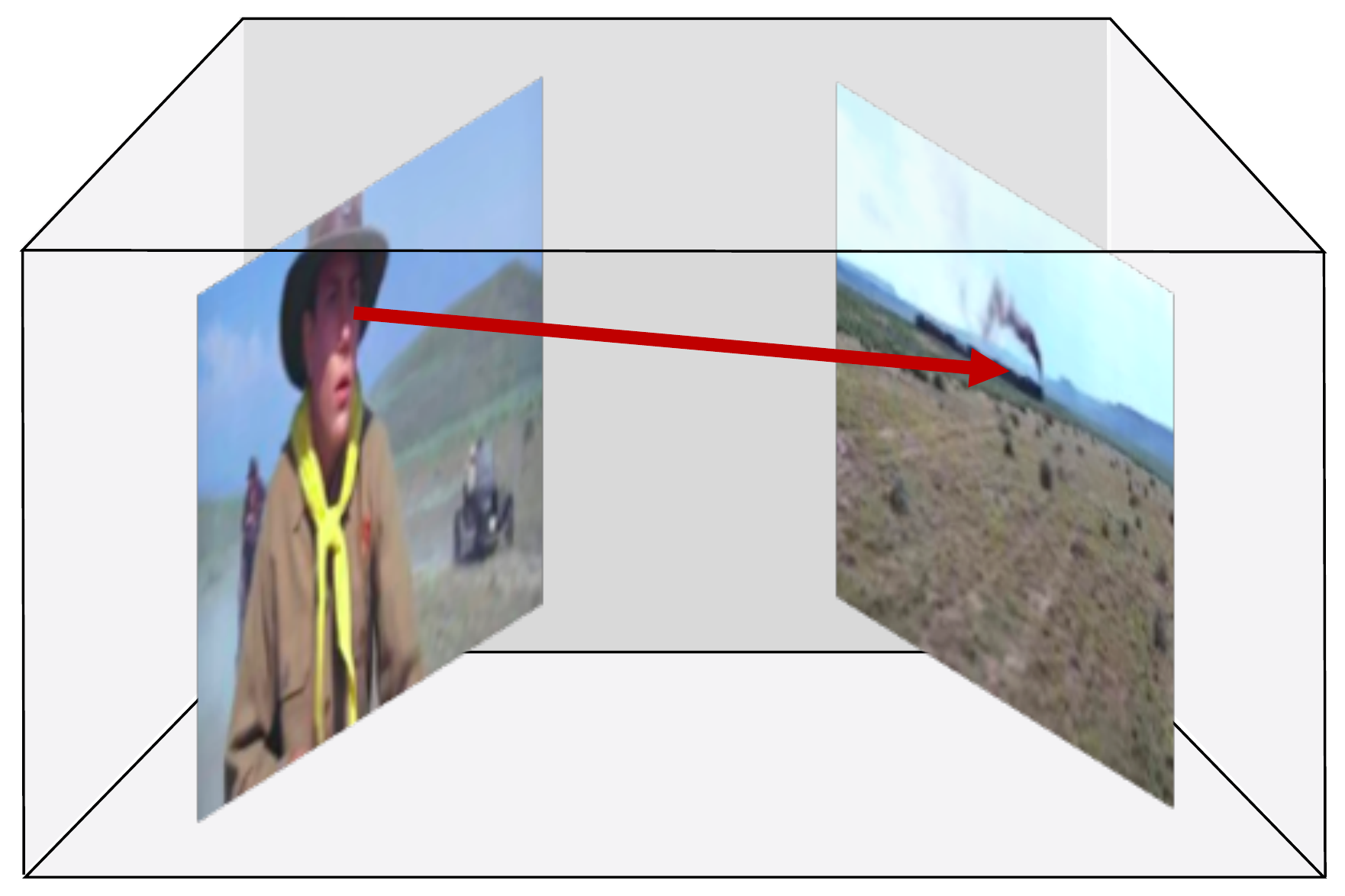}

\caption{We introduce a method for following people's gaze across multiple views of the scene. Given one view with a person in it (left) and a second view (right), we predict  the location of gaze in the second view (right). }
\vspace{-0.5cm}
\label{fig:pool_figure}
\end{figure}

Following the gaze of people is crucial for understanding their actions, activities and motivations. There are many applications that would benefit by the capability to follow gaze in unconstrained scenes e.g., whether the pedestrian has seen the car in the crosswalk \cite{kooij2014context}, if students are paying attention, and understanding social interactions~\cite{vascon2014game,hoai2014talking,chakraborty20133d}. In this paper, we scale up gaze following \cite{recasens2015they} to work across multiple views of the same scene. For example, the scenario in Figure \ref{fig:pool_figure} seems clear: the character is looking at the smoke. Although the person and the smoke are not in the same view, we are able to easily follow their gaze across views and understand that he is looking at the smoke. Solving this task requires a rich understanding of the situation. 

To perform the gaze following task across views, we need to solve three  problems. First, we need to reason about the geometric relationship of the camera pose between the two views. In our example, we need to understand that the two images have been taken with cameras pointing in opposite directions. Second, we need to estimate the head pose of the target person. We want to understand, within the first frame, the direction where  the person is looking. We predict the direction where that person is looking, indicated by the red arrow. Finally, we need to find salient objects in the second view. These will be potential gaze solutions of our problem, since people are known to look at salient locations. To complete our example, the smoke in the second view is an example of a salient spot in the image. With these three ingredients we are able to project the gaze direction from the first view to the second view and find a salient object that intersects the gaze direction. 
 
 \begin{figure*}[t!]
\includegraphics[width=\textwidth]{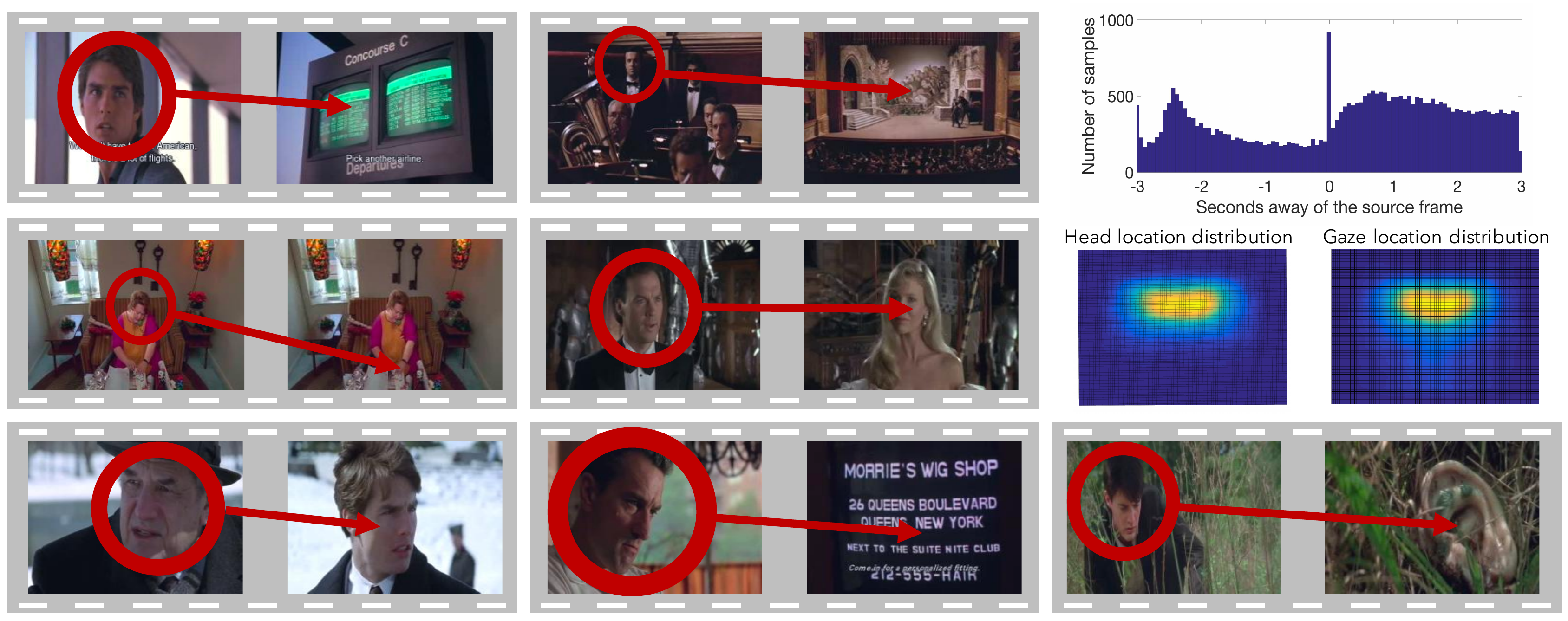}

\caption{\textbf{VideoGaze Dataset:} We present a novel large-scale dataset for gaze-following across multiple views. The histogram at the top right shows the distribution of the time between when a person appears in the clip and the object he is looking at occurs. The heat maps below it show the probability density for the head and gaze location respectively. We also show some annotated examples from the dataset.}

\label{fig:dataset}
\vspace{-0.6cm}
\end{figure*}

We introduce an approach for multi-view gaze following. Given two images taken from different views of a given scene, our method follows the gaze of the people in the scene, even across views. In particular, this system can be applicable to videos, where different frames can be multiple views of the same scenes. Our approach to multi-view gaze following splits the problem into three main modules, each solving a particular task. Although we are dividing the problem into small sub-problems, we are training our system end-to-end providing only gaze labels. Although the sub-modules do not have specific supervision, they learn to automatically solve the sub problems. To both train and benchmark our system, we introduce a new dataset, VideoGaze, for multi-view gaze following using videos \cite{tapaswi2015movieqa}. We annotated $47,456$ people in videos with head, eyes and gaze labels. 

There are three main contributions of this paper. First, we introduce the new problem of following gaze across views. Second, we release a large-scale dataset for both training and evaluation on this task. Third, we present a novel network architecture that leverages the geometry of the scene to tackle this problem. The remainder of this paper details these contributions. In Section 2 we explore related work. In Section 3 we  present our new dataset containing gaze annotation in movies. In Section 4, we describe the model in detail. In the final section, we evaluate the model and provide sample results.


\section{Related Work}

This paper builds upon a previous gaze-following model for static images \cite{recasens2015they}. However, the previous work focuses only on cases where a person, within the image, is looking at another object in the same image. In this work, we remove this restriction and extend gaze following to cases where a person may be looking outside the current view. The model proposed in this paper deals with the situation where the person is looking at another view of the scene.

\textbf{Gaze-Following in video:} Although our model is defined for general scenes, one of its main applications is gaze-following in video. The previous work done for gaze-following in video deals with very restricted settings. Most notably~\cite{marin2011here,marin2014detecting} tackles the problem of detecting people looking at each other in video, by using their head pose and location inside the frame. Although our model can be used with this goal, it is applicable to a wide variety of settings: it can predict gaze when it is located elsewhere in the image (not only on humans) or future/past frame of the video. Mukherjee and Robertson \cite{video_rgbd} use RGB-D images to predict gaze in images and videos. They estimate the head-pose of the person using the multi-modal RGB-D data, and finally they regress the gaze location with a second system. Although the output of their system is gaze location, our model does not need multi-modal data and it is able to deal with gaze location in a different view. 
Extensive work has been done on human interaction and social prediction on both images on video involving gaze \cite{vascon2014game,hoai2014talking,chakraborty20133d}. Some of this work is focused on ego-centric camera data, such as in ~\cite{fathi2012learning,fathi2012social}. Furthermore, \cite{primary_gaze,soo2015social} predicts social saliency, that is, the region that attracts attentions of a group of people in the image. Finally, \cite{chakraborty20133d} estimates the 3D location and pose of the people, which is used to predict social interaction. Although their goal is completely different, we also model the scene with explicit 3D and use it to predict gaze.

\begin{figure*}[t]
\includegraphics[width=\textwidth]{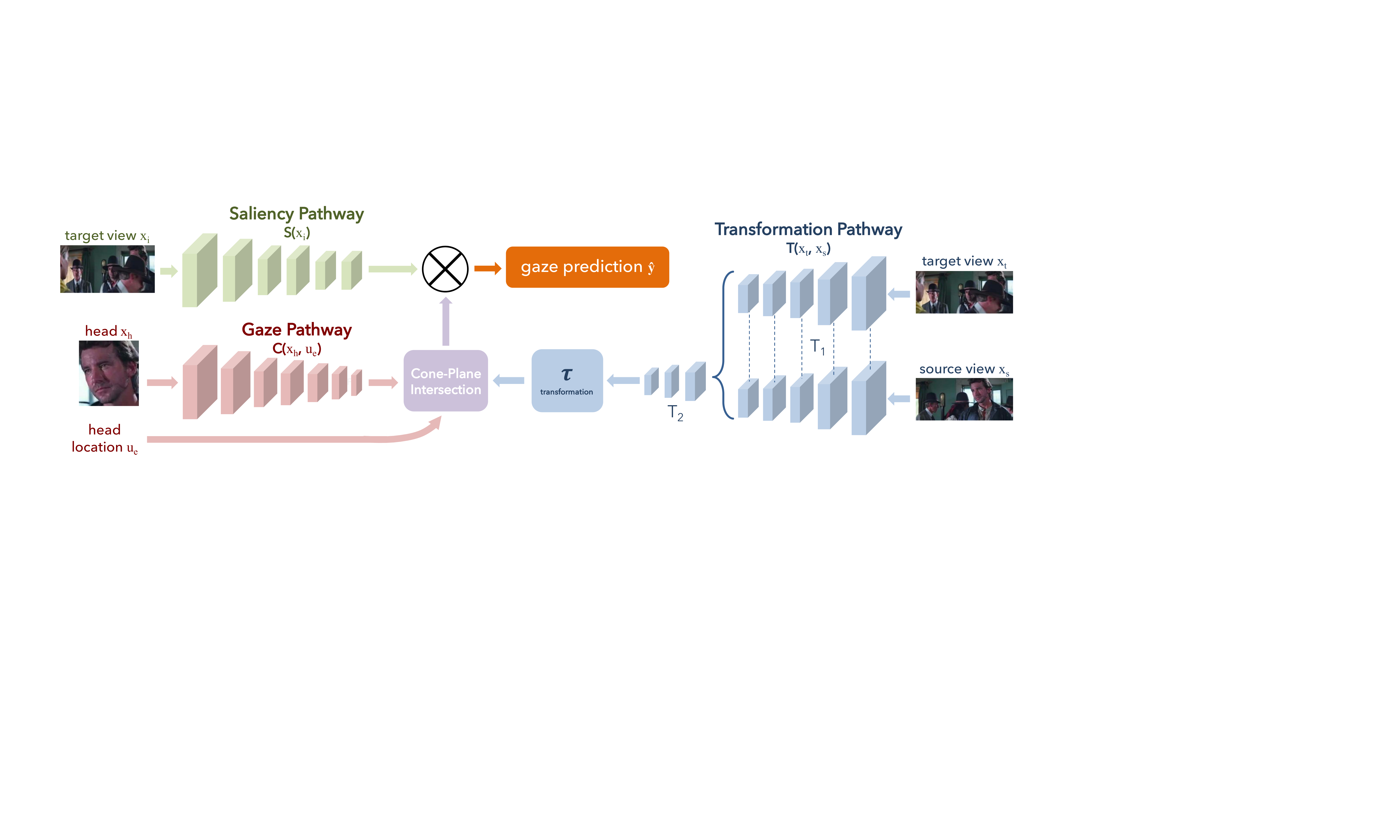}

\caption{\textbf{Network Architecture:} Our model has three pathways. The saliency pathway (top left) finds salient spots on the target view. The gaze pathway (bottom left) computes the parameters of the cone coming out from the person's face. The transformation pathway (right) estimates the geometric relationship between views. The output is the gaze location density.}
\vspace{-0.5cm}

\label{fig:cnn_structure}
\end{figure*}

\textbf{Deep Learning with Geometry:} Unlike \cite{recasens2015they}, we use parametrized geometry transformations that help the model to deal with the underlying geometry of the world. Neural networks have already been used to model transformations, such as in \cite{hinton2011transforming,hinton1981parallel}. Our work is also related to Spatial Transformers Networks \cite{jaderberg2015spatial}, where a localization module generates the parameters of an affine transformation and warps the representation with bilinear interpolation. In this work, our model generates parameters of a 3D affine transformation, but the transformation is applied analytically without warping, which may be more stable.
\cite{rezende2016unsupervised,detone2016deep} used 2D images to learn the underlying 3D structure. Similarly, we expect our model to learn the underlying 3D structure of the frame composition only using 2D images. Finally, \cite{handa2016gvnn} provide efficient implementations for adding geometric transformations to convolutional neural networks. 

\textbf{Saliency:} Although related, gaze-following and free-viewing saliency refer to different problems. In gaze-following, we predict the location of the gaze of an observer in the scene, while in saliency we predict the fixations of an external observer free-viewing the image. Some authors have used gaze to improve saliency prediction, such as in~\cite{parks2014augmented}. Furthermore, \cite{bylinskii2016should} showed how gaze prediction can improve state-of-the-art saliency models. Although our approach is not intended to solve video saliency neither is using video as input, we believe it is worth mentioning some works learning saliency for videos such as \cite{li2009dataset,xia2010novel,li2007fast}.

\section{VideoGaze  Dataset}
\label{sec:dataset}

We introduce \dataset, a large scale dataset containing the location where film characters are looking in movies. \dataset\ contains $47,456$ annotations from $140$ movies. To build the dataset we used videos from the MovieQA dataset \cite{tapaswi2015movieqa}, which we consider a representative selection of movies.
Each sample of the dataset consists of a pair of frames (or views). The first frame of the pair contains the character whose gaze is annotated. Eye location and a head bounding box for the character are provided. The second frame contains the location that character is looking at the time, which can occur temporally after or before the first frame. 

To annotate the dataset, we used Amazon's Mechanical Turk (AMT). We developed an online tool to annotate gaze in videos where the worker is able to first locate the head of the character and then scan through the video to find the location of the object the character is looking at. We also provided options to indicate that the gazed object never appears in the clip or that the head of the character is not visible in the scene. The initial person bounding boxes have been obtained using person detectors from \cite{renNIPS15fasterrcnn}. For quality control, we included samples with known ground truth. We discarded workers that provided poor quality on their annotations. 

We split our data into training set and test set. We use all the annotations from $20$ movies as the testing set and the rest of the annotations as training set. Note that we made the train/test split by source movie, not by clip, which prevents overfitting to particular movies.

Our dataset captures various scenarios present in movies. For instance, $72\%$ of characters in the movies are looking at something that appears at some point in the movie. $27\%$ of characters are looking at something which never appears in the scene. We can further unpack this statistic by analysing the difference in time between the moment when a character appears and the moment when the object he is looking at appears. In Fig. \ref{fig:dataset} (top right) we show the histogram of this time distribution. To summarize, $16.89\%$ of the times when the object is in the movie, it is present in the same frame as the person who is looking at it (the peak at $0$ seconds). One example of annotation in the same frame is shown in Figure \ref{fig:dataset}, the first example of the second row. Furthermore, we can observe that the time when an object of gaze appears in the movie relative to the character is biased towards the future. We also show the spatial distribution of head and gaze spatial distribution in Fig. \ref{fig:dataset}.

\begin{figure*}[t!]
\centering
\includegraphics[width=0.96\textwidth]{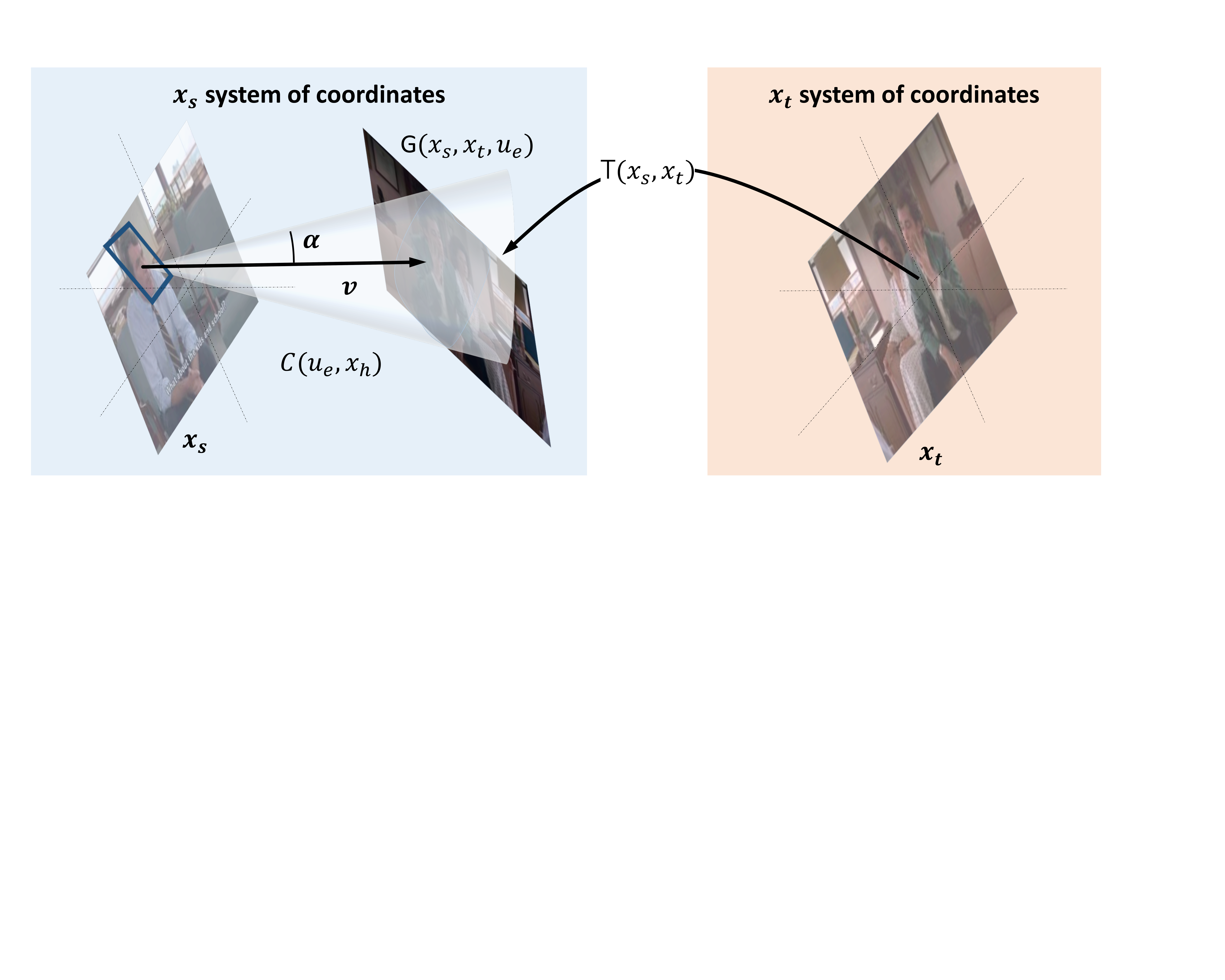}

\caption{\textbf{Transformation and intersection}: The cone pathway computes the cone parameters $v$ and $\alpha$, and the transformation pathway estimates the geometric relation among the original view and the target view. The cone origin is $u_e$ and $x_h$ is indicated with the blue bounding box.}

\label{fig:cone_generator}
\vspace{-0.6cm}
\end{figure*}
\section{Method}

Suppose we have a scene and a person inside the scene. Our goal is to predict where the person is looking, which may possibly be in another view of the scene. Let $x_s$ be the view where the person is located (\textbf{source view}), $x_h$ be an image crop containing only the person's head, and $u_e$ be the coordinates of the eyes of the person within the view $x_s$. Let $x_t$ be the view where we want to predict the gaze location (\textbf{target view}). Given these, we seek to predict the coordinates of the person's gaze $\hat{y}$ in the target view $x_t$. Note $x_t$ may either be the same or different as $x_s$. 

We design a convolutional neural network $F(x_s, x_h, u_e, x_t)$ to predict $\hat{y}$. While we could simply concatenate these inputs and train a network, the internal representation would be entangled and may require large amounts of training data to discover consistent patterns, which is inefficient. Instead, we seek to take advantage of the geometry of the scene to better predict people's gaze.

\subsection{Multi-View Gaze Network}

To follow gaze across views,  the network must be able to solve three sub-problems: (1) estimate the head pose of the person, (2) find the geometric relationship between the view where the person is and the view where the gaze location might be, and (3) find the potential locations in the target view where the person might be looking (salient spots). We design a single model that internally solves each of these sub-problems even though we supervise the network only with the gaze annotations.

We design the network $F$ with this structure in mind:
\begin{align}
    F(x_s, x_h, u_e, x_t) &= S(x_t) \odot G(u_e, x_s, x_t) 
\end{align}
where $S(\cdot)$ and $G(\cdot)$ are decompositions of the original problem. Both $S(\cdot)$ and $G(\cdot)$ produce a positive matrix in  $\mathbb{R}^{k\times k}$ with $k$ being the size of the spatial maps and $\odot$ is the element-wise product.  Although we only supervise $F(\cdot)$, our intention is that $S(\cdot)$ will learn to detect salient objects and $G(\cdot)$ will learn to estimate a mask of all the locations where the person could be looking in $x_t$. We use the element-wise product as an ``and operation'' so that the network predicts people are looking at salient objects that are within their eyesight. 

$S$ is parametrized as a neural network. The structure of $G$ is motivated to leverage the geometry of the scene. We write G as the intersection of the person's gaze cone with a plane representing the target view $x_t$ transformed into the same coordinate frame as $x_s$:
\begin{align}
G(u_e, x_s, x_t) &=  C(u_e,x_h) \cap\tau(T(x_s,x_t))
\end{align}
where $C(u_e, x_s) \in \mathbb{R}^{7}$ estimates the parameters of a cone representing the person's gaze in the original image $x_s$, $T(x_s, x_t) \in \mathbb{R}^{3\times 4}$ estimates the parameters of an affine transformation of the target view, and $\tau$ applies the transformation. $\tau$ is expected to compute the coordinates of $x_t$ in the system of coordinates defined by $x_s$. We illustrate this process in Figure \ref{fig:cone_generator}. 



\subsection{Transformation $\tau$}

We use an affine transformation to geometrically relate the two scenes $x_s$ and $x_t$.  Let $Z$ be the set of coordinates inside the square with corners $(\pm 1, \pm 1,0)$. Suppose the image $x_s$ is located in $Z$ ($x_s$ is resized to have its corners in $(\pm 1, \pm 1,0)$) . Then:
\begin{align}
\tau(T) = Tz \quad \ \forall z \ \in Z
\end{align}
The affine transformation $T$ is computing the geometric relation between both views. To compute the parameters $T$ we used a convolutional neural network. We use $T$ to transform the coordinates of $x_t$ into the coordinate system defined by $x_s$.

\subsection{Cone-Plane Intersection}

Given a cone parametrization of the gaze direction $C$ and a transformed view plane $\tau(T)$, we wish to find the intersection $C \cap \tau(T)$. The intersection is obtained by solving the following equation for $\beta$:
\begin{equation}
\beta^T \Sigma \beta = 0 \ \textrm{ where } \\  \beta = (\beta_1, \beta_2,1)
\label{eqn:intersection}
\end{equation}
where $(\beta_1, \beta_2)$ are coordinates in the system of coordinates defined by $x_t$, and $\Sigma \in \mathbb{R}^{3\times3}$ is a matrix defining the cone-plane intersection as in \cite{cone_generation}. Solving Equation \ref{eqn:intersection} for all $\beta$ gives us the cone-plane intersection, however it is not discrete, which would not provide a gradient for learning. Therefore, we use an approximation to make the intersection soft:
\begin{align}
\label{eqn:ohgod}
C(u_e,x_h) \cap \tau(T(x_s,x_t)) = \sigma( \beta^T \Sigma \beta  )
\end{align}
where $\sigma$ is a sigmoid activation function. To compute the intersection, we calculate Equation \ref{eqn:ohgod} for $\beta_1, \beta_2 \in [-1,1]$.

\textbf{Deriving $\Sigma$:} We summarize the derivation of $\Sigma$ here, but we refer readers to the supplemental materials for complete details.
A cone in the space with origin $u_e$, matrix $M$ and vector $v$ can be parametrized as all the points $p$ such that:
\begin{equation}
\label{eqn:cone_M}
    (p-u_e)^TM(p-u_e)  = 0
\end{equation}
where $M$ can be computed as $M = v^tv-\alpha I$ where I is the identity matrix. 
Since we assume $x_s$ is in $Z$, the square with corners $(\pm 1, \pm 1,0)$, we can parametrize $\tau(T)$ as a plane with unit vectors $v_1 = Re_1$ and $v_2 = Re_2$ where $e_i$ is the orthonormal basis and $R$ is the linear part of the affine transformation. All the points of the plane can be written as $p = t + \beta_1 v_1 +  \beta_2 v_2$. We derive $\Sigma$ by substituting $p$ into Equation \ref{eqn:cone_M}.

\subsection{Pathways}

We estimate the parameters of the saliency map $S$, the cone $C$, and the transformation $T$ using convolutional neural networks.

\textbf{Saliency Pathway:}
The saliency pathway uses the target view $x_t$ to generate a spatial map $S(x_t)$. We used a $6$-layer convolutional neural network to generate the spatial map from the input image. The five initial convolutional layers follow the structure of AlexNet introduced by~\cite{krizhevsky2012imagenet}. The last convolutional layer uses a $1 \times 1$ kernel to merge the $256$ channels in a simple $1 \times k \times k$ map.

\textbf{Cone Pathway:}
The cone pathway generates a cone parametrization from a close-up image of the head $x_h$ and the eyes $u_e$. We set the origin of the cone at the head of the person $u_e$ and let a convolutional neural network generate $v \in \mathbb{R}^3$, the direction of the cone and $\alpha \in \mathbb{R}$, its aperture. Figure \ref{fig:cone_generator} shows an schematic example of the cone generation. Additionally to the cone direction vector and the aperture value, we also estimate a radius $r \in \mathbb{R}$ which is used to put a virtual ball around the person's head to avoid degenerated solutions where people look at themselves.

\textbf{Transformation Pathway:}
The transformation pathway has two stages. We define $T_1$, a $5$-layer convolutional neural network following the structure defined in~\cite{krizhevsky2012imagenet}. $T_1$ is applied separately to both the source view $x_s$ and the target view $x_t$. We define $T_2$ which is composed by one convolutional layer and three fully connected layers reducing the dimensionality of the representation to a low dimension representation. The output of the pathway is computed as: $T(x_s,x_t) = T_2(T_1(x_s),T_1(x_t))$. We used \cite{handa2016gvnn} to compute the transformation matrix from output parameters.

\tabcolsep=0.11cm

\begin{table*}[t!]
\centering
\begin{subtable}[b]{0.48\linewidth}
\centering
\scalebox{1}{
\begin{tabular}{ l | c c c c}
Model                             & AUC                 & Dist & AUC $\ge 1s$ & Dist $\ge 1s$ \\ \hline
Static Gaze \cite{recasens2015they}  & $0.770$ & $0.296$         &       $0.763$ &   $0.304$           \\ 
Judd \cite{judd2009learning}         & $0.810$ & $0.343$         &         $0.825$   & $0.335$         \\  
Fixed bias                           & $0.656$ & $0.346$        &                    $0.646$  & $0.352$         \\ 
Center                               & $0.514$ & $0.249$        &                    $0.510$  & $0.253$         \\ 
Random                               & $0.600$ & $0.470$        & $0.598$   &$0.471$              \\ 
\hline
Our (vertical axis rot)                  & $\mathbf{0.844}$    & $\mathbf{0.209}$         &  $\mathbf{0.866}$ & $\mathbf{0.187}$  
\end{tabular}

}
\caption{Baselines}
\label{table:gaze_prediction}

\end{subtable}\hspace{0.02\linewidth}\begin{subtable}[b]{0.48\linewidth}
\centering
\footnotesize
\scalebox{1}{
\begin{tabular}{ l|c c c c }
Model                             & AUC                 & Dist & AUC $\ge 1s$ & Dist $\ge 1s$ \\ \hline

No image                          & $0.810$ &    $0.216$     &    $0.849$    & $0.184$      \\ 
No cone layer                     & $0.779$ & $0.239$        &      $0.809$   & $0.212$      \\ 
No head                           & $0.841$ & $0.233$  &    $0.874$   & $0.192$     \\ 
Identity        &$0.825$ & $0.223$        & $0.856$   & $0.189$          \\ 
Translation only                       & $0.829$ & $0.213$        &       $0.861$   & $0.183$   \\ 
Rotation only                    &$0.803$ & $0.228$        & $0.841$   & $0.192$          \\ 
$3$-axis rotation                 & $0.828$ & $0.216$        &      $0.862$   & $0.182$      \\ 
Vertical axis rotation            & $0.844$    & $0.209$         &  $0.866$ & $0.187$   \\ 
\end{tabular}
}
\vspace*{-2mm}
\caption{Model Analysis}

\label{table:model_diagnostic}
\end{subtable}
\vspace*{-3mm}
\caption{\textbf{Evaluation:} In table (a) we compare our performance with the baselines. In table (b) we analyse the performance of the different ablations and variations of our model. \textit{AUC} stands for Area Under the Curve and it is computed as the to the area under the ROC curve. Higher is better. \textit{Dist.} is computed as the $L_2$ distance to the ground truth location. Lower is better. We also compute both metrics for frames where the annotation is one second or more away from the original frame. }
\label{tables:evaluation}
\vspace*{-2mm}
\end{table*}

\textbf{Discussion:} We constrain each pathway to learn different aspects of the problem by providing each pathway only a subset of the inputs.
The saliency pathway only has access to the target view $x_t$, which is insufficient to solve the full problem. Instead, we expect it to find salient objects in the target view $x_t$. Likewise, the transformation pathway has access to both $x_s$ and $x_t$, and the transformation will be later used to project the gaze cone. We expect it to compute a transformation that geometrically relates $x_s$ and $x_f$.
We expect each of the pathways to learn to solve its particular subproblem to then get geometrically combined to generate the final output. Since every step is differentiable, it can be trained end-to-end without intermediate supervision.

\subsection{Learning}
Gaze-following is a multimodal problem  \cite{recasens2015they}. For this reason, we choose to estimate a probability heat map for prediction instead of regressing a single gaze location. We use a shifted grids spatial loss from \cite{recasens2015they}, which helped with localization. We created $5$ different classification grids with side length of $5$, and finally combine the predictions. Each of the grids is spatially shifted towards one direction of the image, creating $5$ overlapping but different classification problems. Using this learning procedure, the model can learn to generate higher precision results by solving multiple low-precision problems simultaneously. We found that using the shifted grids output slightly improves performance over the element-wise product output.

\subsection{Inference}

The predictor will produce a matrix in $\mathbb{R}^{15 \times 15}$ (the shifted grids procedure produces an output map $3$ times larger than the side length chose). This map $A$ can be interpreted as a density where the person is looking. To infer the gaze location $\hat{y}$ in the target frame $x_t$, we simply find the mode of this density $\hat{y} = \argmax_{i,j} A_{ij} $.

\subsection{Looking Outside The Frame}
\label{subsect:outside}
Our method aims to solve the problem of multi-view gaze following. However, in applications such as gaze following in video, we might need to estimate whether two frames are taken from the same scene or from different scenes. So far, our transformation pathway is able to estimate the geometric relation among views of the same scene: for views belonging to the same scene we are able to correctly follow gaze and predict where the person is looking in this view. In this section, we extend our model to predict whether two views are coming or not from the same scene. 

In the extended model, our transformation pathway additionally computes a confidence value $\gamma \in [0,1]$ indicating whether two views are part of the same scene or not. The cone generator will scale the cone projection with $\gamma$. If the views are part the same scene, $\gamma \approx 1$ and the cone projection remains intact. However, if the views are part of different scenes then $\gamma \approx 0$ and the cone projection will be ignored. $\gamma$ is directly supervised with a Cross-Entropy Loss. The final loss of the training is a linear combination between the gaze prediction and the same-scene classification task. An extra class is added to the final prediction to account for samples without gaze label. Our findings indicate that there is a trade-off in performance between both criterion. In this experimental section we will quantify the performance of the extended model. 

\subsection{Implementation Details}
We implemented our model using Torch. In our experiments we use $k=13$; the output of both the saliency pathway and the cone generator is a $13 \times 13$ spatial map. We initialize the convolutional networks in three pathways with Imagenet-CNN \cite{krizhevsky2012imagenet}. The cone pathway has three fully connected layers of sizes $500$, $200$ and $5$ to generate the cone parametrization. The common part of the transformation pathway, $T_2$, has one convolutional layer with a $1 \times 1$ kernel and $100$ output channels, followed by one $2 \times 2$ max pooling layer and three fully connected layers of $200$, $100$ and the parameter size of the transformation. For training, we augment data by flipping $x_t$ and $x_s$ and their annotations.

\section{Experiments}
\label{sec:experiments}
\begin{figure*}[t]
\includegraphics[width=\textwidth]{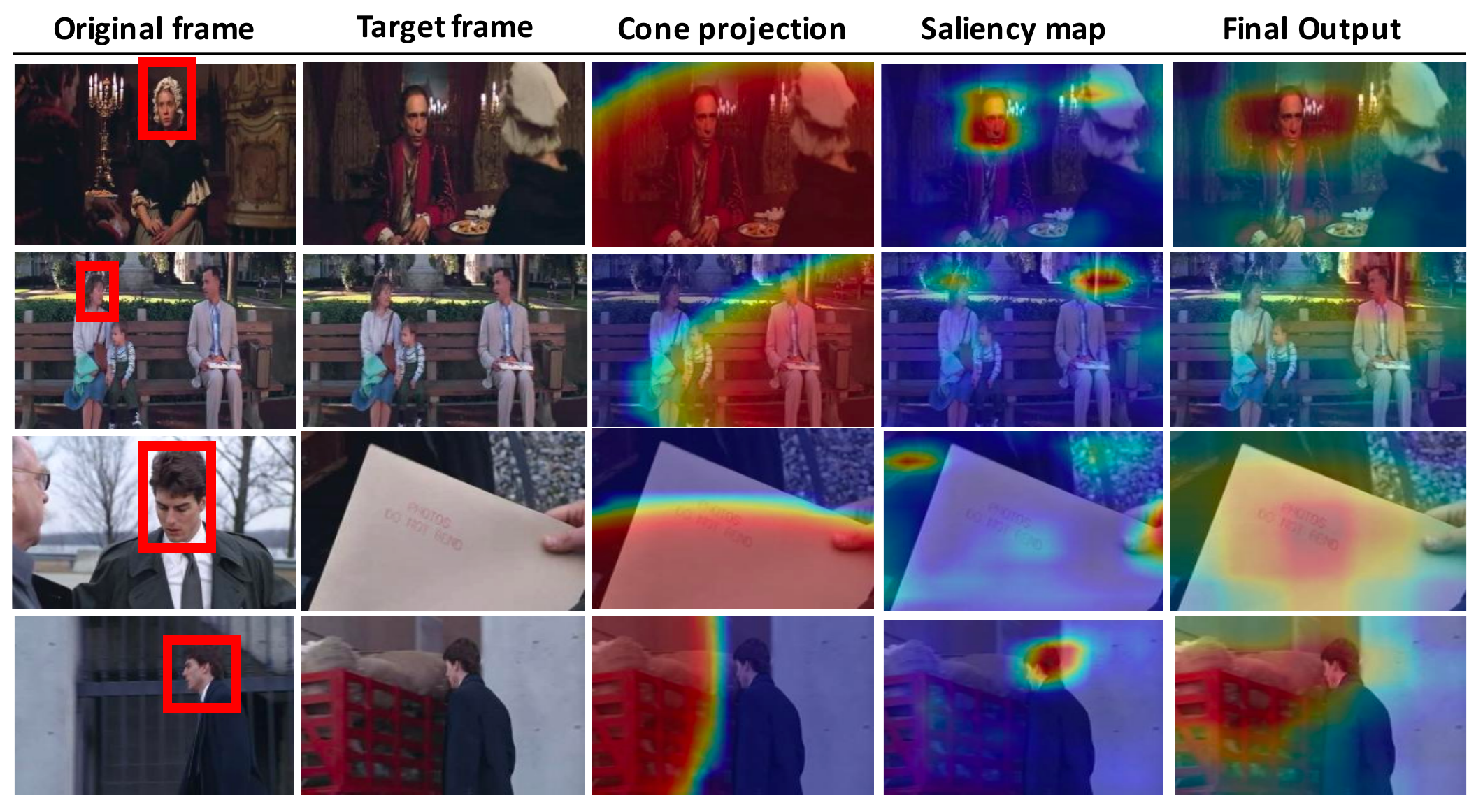}
\vspace*{-8mm}
\caption{\textbf{Internal visualizations:} We show examples of the output for the different pathways of our network. The cone projection shows the final output of the cone-plane intersection module. The saliency map shows the output of the saliency pathway. The final output show the predicted gaze location distribution.}
\vspace*{-5mm}
\label{fig:internal}
\end{figure*}

\subsection{Evaluation Procedure}
To evaluate our model we conducted quantitative and qualitative analyses using our held out dataset. Similar to \cite{PASCAL}, we provide bounding boxes for the heads of the persons. The bounding boxes are part of the dataset and have been collected using Amazon's Mechanical Turk. This makes the evaluation focused on the gaze following task. In the supplemental materials we provide some examples of our system working with head bounding boxes computed with an automatic head detector. 

We use AUC and $L_2$ distances as our evaluation metrics. AUC refers to Area Under the Curve, a measure typically used to compare predicted distributions to samples. The predicted heat map is used as a confidence to build a ROC curve. We used \cite{judd2009learning} to compute the AUC metric. Furthermore, we used $L_2$ metric, which is computed as the euclidean error between the predicted point and the ground truth annotation. For comparison purposes, we assume the images are normalized to having sides of length $1$ unit. 

Although the main evaluation is done through the full test set, a second evaluation is performed over the samples where the source frame is more than $1$ second away from the target view. This evaluation is intended to show the performance of our model is scenarios where the rotation and translation are larger. 

Previous work in gaze following in video cannot be applicable to our experiment because of its particular contains (only predicting social interaction or using multi-model data). We compare our method to several baselines described below. For methods producing a single location as output, the output heatmap is a Gaussian distribution centered in the output location. 

\textbf{Random}: The prediction is a random location in the image. 
\textbf{Center}: The prediction is always the center of the image. 
\textbf{Fixed bias}: The head location is quantized in a $13 \times 13$ grid and the training set is used to compute the average output location per each head location. 
\textbf{Saliency}: The output heatmap is the saliency prediction for $x_t$. \cite{judd2009learning} is used to compute the saliency map. The output point is computed as the mode of the saliency output distribution.
\textbf{Static Gaze}: \cite{recasens2015they} is used to compute the gaze prediction. Since it is a method for static images, the head image and the head location provided are from the source view but the image provided is the target view. 

Additionally, we performed an analysis on the components of our model. With this analysis, we aim to understand the contribution of each of the parts to performance as well as suggest that all of them are needed.

\textbf{Translation only}: The affine transformation is a translation. 
\textbf{Rotation only}: The affine transformation is a rotation in the $x$-axis.
\textbf{Identity}:  The affine transformation is the identity. 
\textbf{No head}: The saliency pathway is used to generate the output.
\textbf{No image}: The gaze pathway combined with the transformation pathway are used to generate the output. The saliency pathway is ignored.
\textbf{No cone layer}: The cone layer is substituted with $3$ fully connected layers going from all the parameters to the $13 \times 13$ spatial map.
\textbf{3 axis rotation / translation}: The affine transformation is a $3$ axis rotation combined with a translation.  
\textbf{Vertical axis rotation}: The affine transformation is a rotation in the vertical axis combined with a translation.

\begin{figure}[t]
\includegraphics[width=0.48\textwidth]{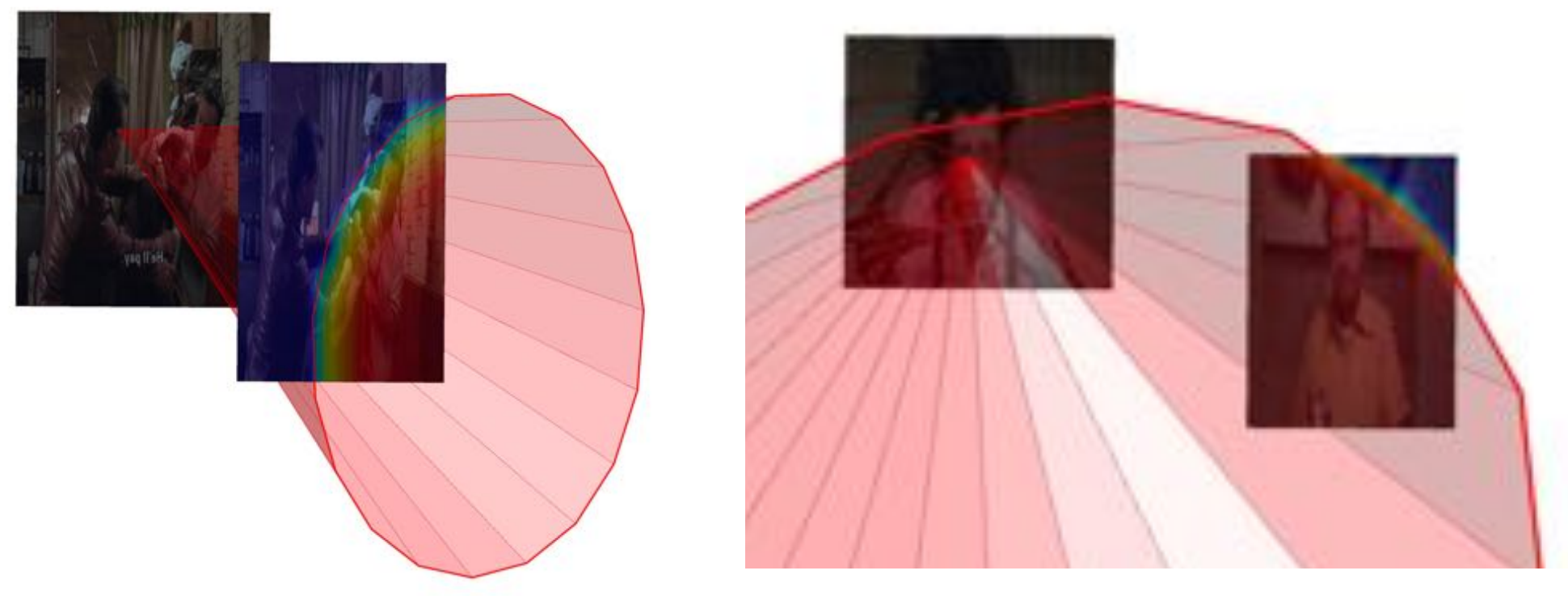}

\caption{\textbf{Cone visualizations:} We show examples of the output of the cone-plane intersection, plotting the images in the relative location estimated by $T$. Best viewed on screen.}
\vspace{-0.5cm}
\label{fig:internal_cone}
\end{figure}

\begin{figure}[t!]
\includegraphics[width=0.48\textwidth]{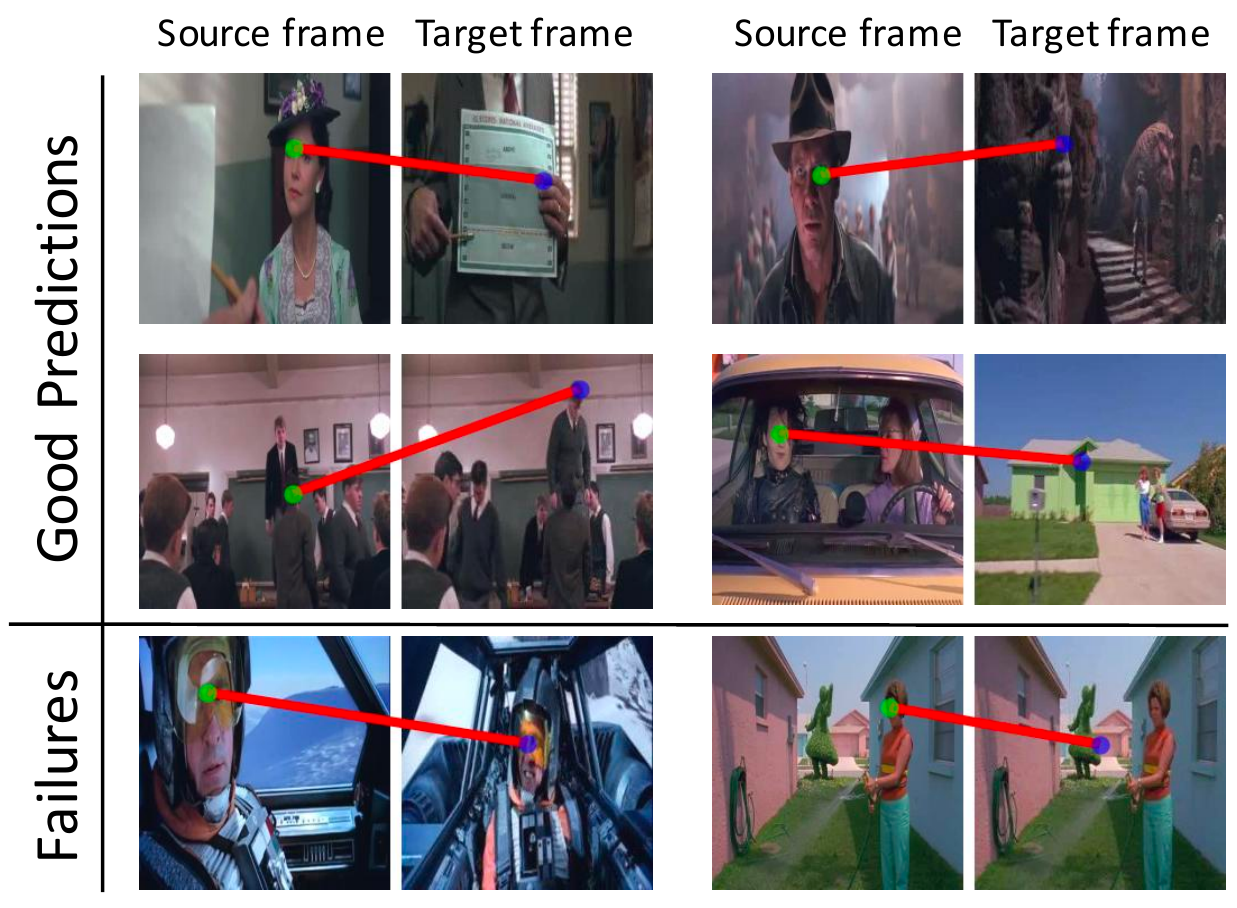}

\caption{\textbf{Output examples:} We show the output of our method in the test set, including two common failures: within-image 3D reasoning and missing context. Green dot indicates head location and blue dot gaze prediction.}
\vspace{-0.6cm}
\label{fig:simple_results}
\end{figure}

\subsection{Results}

Table~\ref{tables:evaluation} shows the performance of all the models and variations in both metrics. Our model has a performance of $0.844$ in AUC and $0.209$ in $L_2$ in the full testing set and $0.866$ / $0.187$ in the reduced test set of only images temporally far ($1$ second or more) from the source. Our model's performance is significantly better than all the baselines. Interestingly, the performance of most of the models increase when both views are significantly different. This is due to the fact that, if the views are different, the second view is likely to be more focused on the object of gaze of the person. Note that the static model is the only one that worsen in the reduced test set. This is due to its design to handle situations where the object is in the same view.

Our analysis show that our model outperforms all possible combinations of models and affine transformations. It is a natural outcome that the best model is restricted only to rotations over the vertical axis and translation, given that these are typical movements for cameras. Interestingly, the performance of the ablations is closer in the reduced testing set. As previously mentioned, in this scenario the saliency pathway is more important than the gaze pathway.

Figure \ref{fig:simple_results} shows the output of our model for some test samples. We present four cases where our prediction is correct and two with failures. In the first failure case (left), the model predicts that the two aviators are looking at each other. This is not a possible situation given that they are in different planes, but the model is not given enough information to understand the context. In the second failure case (right), the person is predicted to be looking at the house behind her. It is clear that this situation is impossible, but our model does not have 3D reasoning within the images, it only estimates the 3D relationship between images.

In Figure~\ref{fig:internal} we show the output of the internal pathways of our model. Further, in Figure~\ref{fig:internal_cone} we show two examples of the estimated geometric relationship among views. Both figures suggest that our network has internally learned to solve the sub-problems we intended it to solve, in addition to solving the overall gaze following problem. The network is able to estimate the geometrical relationship among frames (see Figure~\ref{fig:internal_cone}) along with estimating the gaze direction from the source view and predicting the salient regions in the target view. 

\subsection{Looking outside the scene}
Here, we evaluate the extension of our model designed to detect large scene changes. We augmented our dataset with frames from the same videos but different scenes, to include examples of views from different scenes. Using this extra annotation, we trained the model and evaluated its ability to detect scene changes. We used average precision (AP) to evaluate the task. AP is commonly used to evaluate detection tasks, and is computed as the area below the precision-recall curve. In our test set, chance is $0.5$. Our extended model has a mean average precision of $0.877$ in detecting scene changes, demonstrating that our performance is significantly above chance. Examples of the extended model output are shown in the supplemental material.

\subsection{Time analysis}

\begin{figure}[t]
\includegraphics[width=0.48\textwidth]{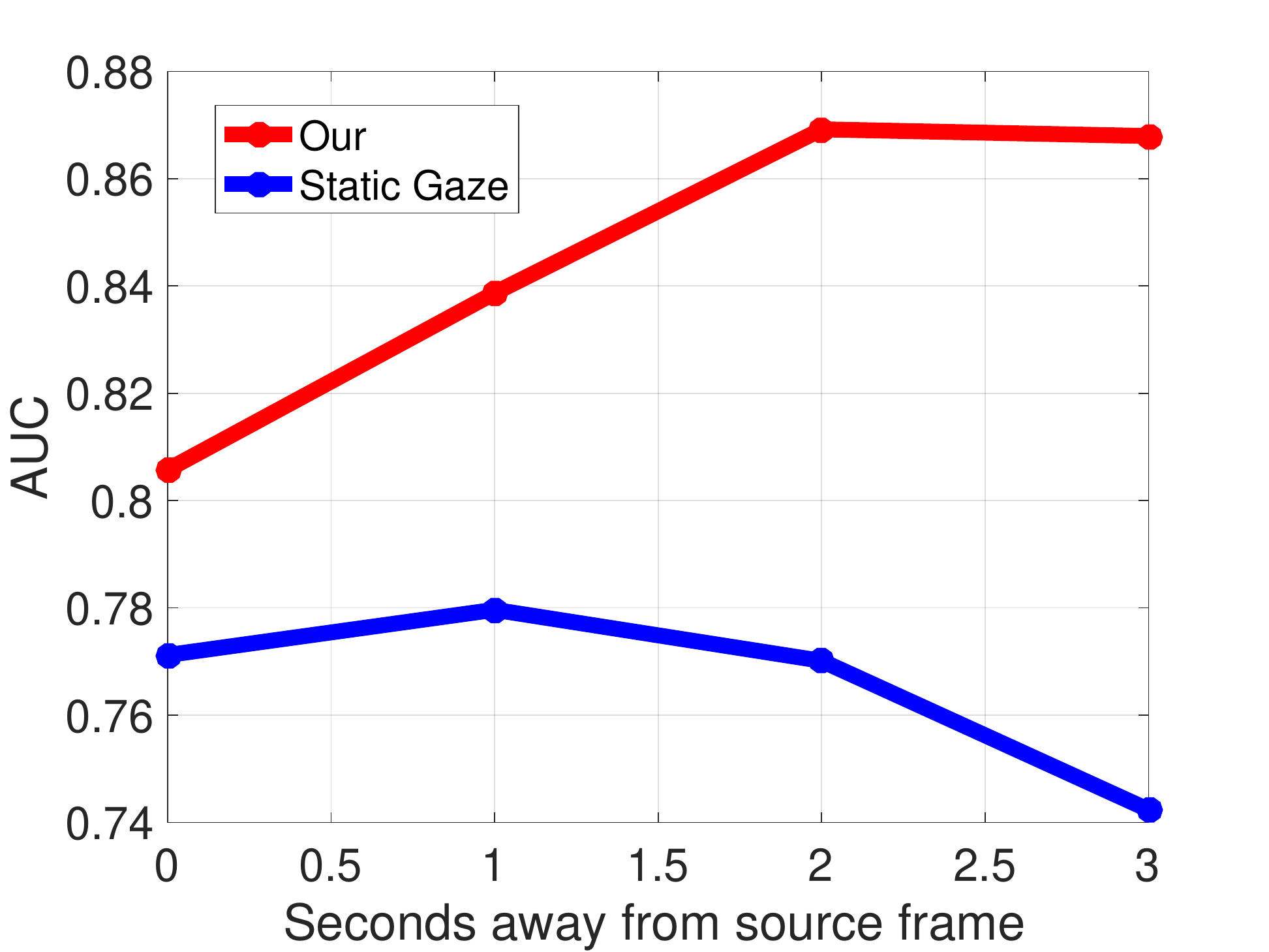}

\caption{\textbf{Time-performance representation:} We plot performance versus the time distance between the target frame and the source frame, which correlates with the size of the transformation. Our model is able to perform well even in situations where the two frames are very different. \cite{recasens2015they} performs worse in examples where the images are different.}
\vspace{-0.4cm}
\label{fig:time_figure}
\end{figure}

To evaluate the performance of our model on different scenarios, in Fig.~\ref{fig:time_figure} we plot the performance of our model when varying the amount of time passage between source and target frame. We also plot the performance of \cite{recasens2015they}. Our model performs better with frames farther in time from the target. However, we can observe how the static gaze model decreases its performance when the views are more different, by its construction of dealing with static images. This shows our model works well in cases when the views are very different.

\section{Conclusions}
We present a novel method for gaze following across views . Given two views, we are able to follow the gaze of a person in a source view to the target view even when the views are quite different. We split our model in different pathways which automatically learn to solve the three main sub problems involved in the task. We take advantage of the geometry of the scene to better predict people's gaze. We also introduce a new dataset where we benchmark our model and show that our method over performs the baselines and produces meaningful outputs. We hope that our dataset will attract the community attention to the problem.

\section{Acknowledgments}
We thank Zoya Bylinskii for proof-reading. Funding for this research
was partially supported by the Obra Social “la Caixa” Fellowship for Post-Graduate Studies to AR, a Google PhD Fellowship to CV and Samsung.

{\small
\bibliographystyle{ieee}
\bibliography{egbib}
}

\end{document}